\documentclass{article}
\usepackage{spconf,amsmath,graphicx,hyperref}
\usepackage{enumitem, bm, amssymb, arydshln, booktabs, multirow, cite, setspace, etoolbox, xcolor, marvosym}
\usepackage{float,enumitem}

\title{Visual Puns from Idioms: An Iterative LLM-T2IM-MLLM Framework}
\name{
	Kelaiti Xiao\textsuperscript{1,2}\quad
	Liang Yang\textsuperscript{1}\qquad
	Dongyu Zhang\textsuperscript{1}\qquad
    PAERHATI Tulajiang\textsuperscript{1,2}\quad
	Hongfei Lin\textsuperscript{1*}
}

\address{
	$^{1}$Dalian University of Technology, Dalian, China\\
	$^{2}$Xinjiang Normal University, Urumqi, China
}

\makeatletter
\patchcmd{\@maketitle}{\vskip 1.5em}{\vskip 1em}{}{}
\patchcmd{\@maketitle}{\@name \\ \@address}{\@name \\[-1em] \@address}{}{}
\patchcmd{\@maketitle}{\vskip 1.5em}{\vskip 1em}{}{}
\makeatother
\begin{document}
	\maketitle
	\begin{abstract}
We study idiom-based visual puns—images that align an idiom’s literal and figurative meanings—and present an iterative framework that coordinates a large language model (LLM), a text-to-image model (T2IM), and a multimodal LLM (MLLM) for automatic generation and evaluation. Given an idiom, the system iteratively (i) generates detailed visual prompts, (ii) synthesizes an image, (iii) infers the idiom from the image, and (iv) refines the prompt until recognition succeeds or a step limit is reached. Using 1{,}000 idioms as inputs, we synthesize a corresponding dataset of visual pun images with paired prompts, enabling benchmarking of both generation and understanding. Experiments across 10 LLMs, 10 MLLMs, and one T2IM (Qwen-Image) show that MLLM choice is the primary performance driver: GPT achieves the highest accuracies, Gemini follows, and the best open-source MLLM (Gemma) is competitive with some closed models. On the LLM side, Claude attains the strongest average performance for prompt generation. Code is available at \url{https://github.com/xkt88/VisualPun}.
	\end{abstract}
    \vspace{-0.5em}
	\begin{keywords}
		Visual Puns, Idiom Understanding, Multimodal Large Language Models, Benchmark dataset\vspace{-0.5em}
	\end{keywords}	
\vspace{-0.5em}
\section{Introduction}
\label{sec:intro}
\vspace{-0.5em}

Visual puns are a sophisticated form of multimodal communication that combines linguistic wordplay with visual representation, requiring simultaneous processing of literal and metaphorical meanings~\cite{visual_pun, ma2025pun2pun}. Creating and interpreting such puns demands resolving semantic ambiguities, aligning visual and linguistic cues, and exercising creative reasoning~\cite{hempelmann2007visual}. These properties make visual puns valuable benchmarks for assessing AI models' creative and interpretive capabilities~\cite{Yuan2025A}. 

\begin{figure}[!t]
  \centering
  \includegraphics[width=\linewidth]{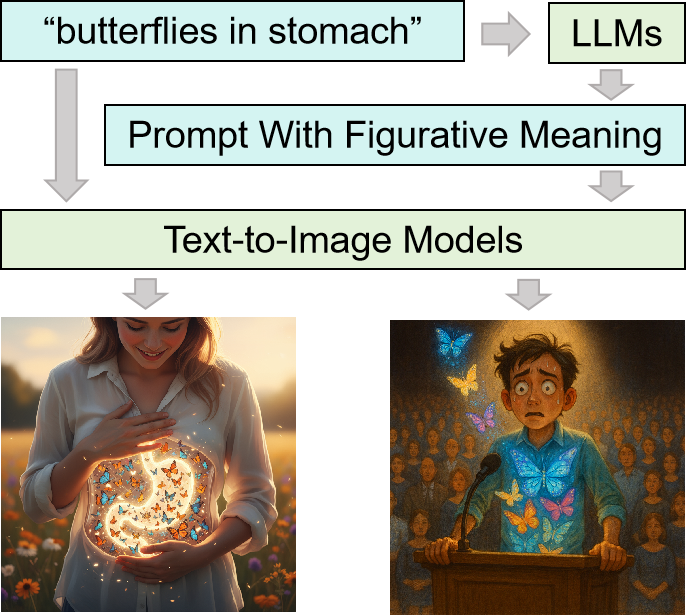}
  \vspace{-2em}
  \caption{
    \small An LLM bridges literal and metaphorical meanings of an idiom, generating prompts that guide a T2IM to create visual puns capturing both nervous feelings and literal butterflies.
  }
  \vspace{-1.5em}
  \label{fig:1}
\end{figure}

Given recent advances in T2IMs\cite{t2i1,t2i2,pragmatic}, a natural direction is to leverage them to automatically synthesize visual-pun datasets at scale. State-of-the-art T2IMs can faithfully render specific scenes when provided with sufficiently detailed prompts~\cite{T2IMs1,T2IMs2,tu-etal-2025-automatic}. However, they lack the deeper linguistic reasoning needed to capture figurative intent, so metaphorical prompts are often misinterpreted~\cite{zhang2024gome}. An effective strategy is to leverage an LLM to decompose an idiom's literal and metaphorical facets and translate them into concrete visual directives for the T2IM, as illustrated in Fig.~\ref{fig:1}. Even with such LLM‑guided prompting, the community still lacks standardized resources and reliable procedures.
\vspace{-0.2em}

Despite rapid progress in text-to-image generation and multimodal modeling~\cite{t2i3,t2i4,t2i5,rewarding}, two gaps remain: (i) there is no large-scale public benchmark specifically targeting idiom-based visual puns with synthesized images~\cite{li2024survey}; existing resources focus on visual metaphors or rebus art and differ in scope~\cite{wordplay,meta,vimeta}; and (ii) one-shot generations are unreliable due to hallucinations and semantic ambiguity~\cite{hall}, motivating iterative self-improvement loops that detect, diagnose, and correct mismatches between intent and output. To address these challenges, we contribute:
\begin{itemize}[leftmargin=*,nosep]
\item An iterative LLM–T2IM–MLLM pipeline that decomposes idioms into literal and figurative cues, generates/refines prompts, synthesizes images, and stops when the idiom is recognized or a step cap is reached.
\item Using 1,000 English idioms, we produce a one-to-one dataset of visual-pun images with paired prompts, enabling benchmarking of generation and understanding.
\item A large-scale evaluation across 10 LLMs and 10 MLLMs showing MLLM choice dominates performance, while T2IM choice matters less with detailed prompts.
\end{itemize}

\begin{figure*}[!t]
    \centering
    \includegraphics[width=\textwidth]{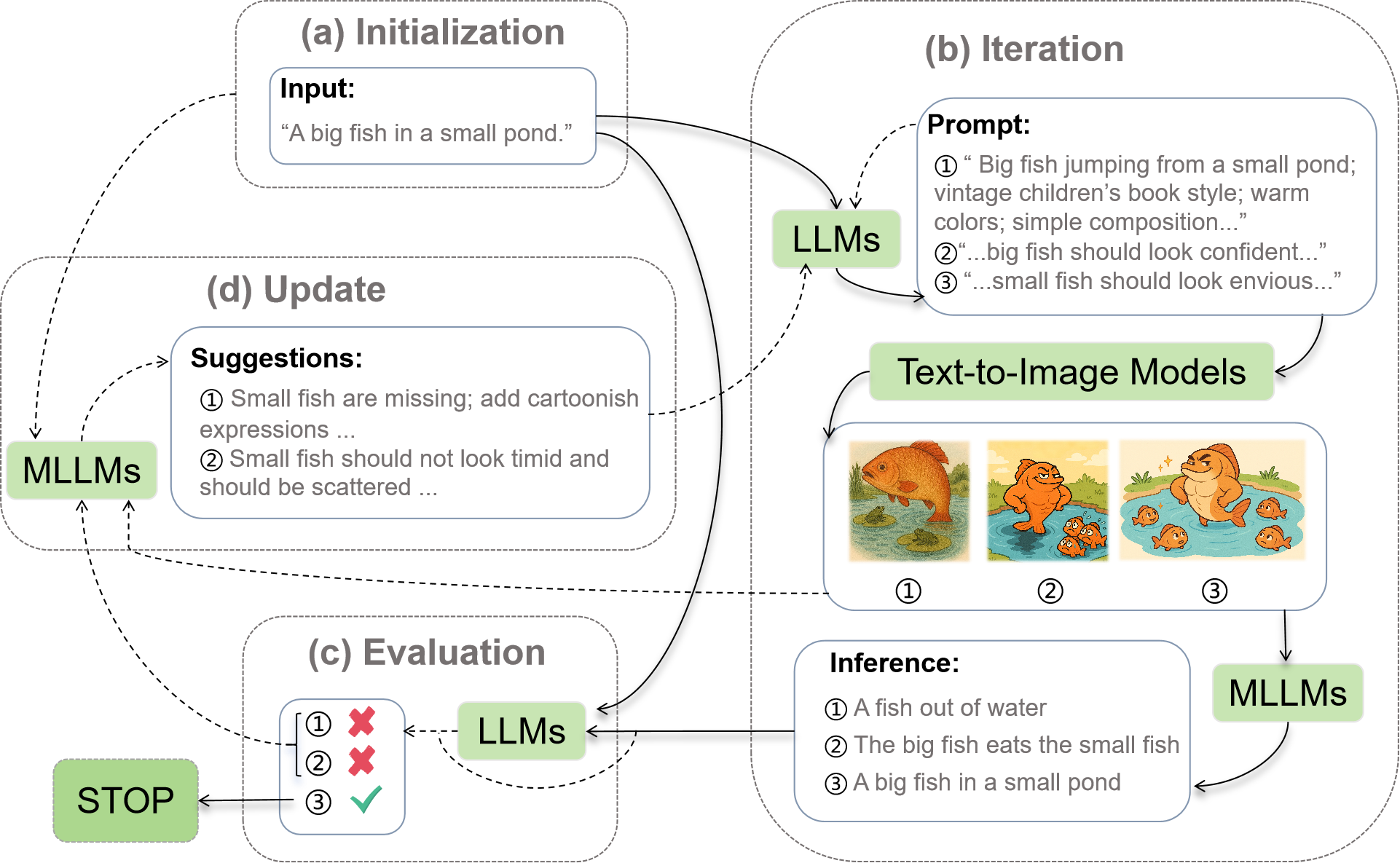}
    \vspace{-2.5em}
    \caption{ \small Iterative pipeline for visual pun generation: the LLM crafts prompts, the T2IM synthesizes images, the MLLM infers the idiom, and the LLM evaluates and updates; solid arrows denote the current iteration, dashed arrows reference the previous iteration}
    \vspace{-1.5em}
    \label{fig:2}
\end{figure*}
      
\vspace{-1.5em}

\section{Methodology}
\label{sec:methodology}
\vspace{-0.5em}
Our pipeline (Fig.~\ref{fig:2}) iterates four modules to render visual puns that convey literal and figurative meanings. Given an idiom $I_{\text{input}}$, an LLM proposes a visual prompt $P_t$ conditioned on prior prompt $P_{t-1}$ and edit suggestions $U_{t-1}$. A T2IM synthesizes an image $G_t$; an MLLM infers the idiom $R_t$ from $G_t$. An LLM judge checks semantic equivalence and issues a control signal; on mismatch, the MLLM supplies edits $U_t$ to refine $P_t$. We stop on a match or after $t{=}5$ iterations.

\vspace{-0.8em}
\subsection{Initialization}
\vspace{-0.4em}
Given an idiom corpus $\mathcal{I}=\{I_1,\dots,I_n\}$, select one target idiom:
\begin{equation}
    I_{\text{input}}=I_j,\quad j\in\{1,\dots,n\}.
\end{equation}
Initialize $P_0=\varepsilon$ (empty prompt) and $U_0=\emptyset$ (no suggestions).

\vspace{-0.8em}
\subsection{Iteration}
\label{sec:iteration}
\vspace{-0.4em}
At iteration $t$:
\begin{align}
    P_t &= \mathrm{LLM}_{\text{prompt}}(I_{\text{input}}, U_{t-1}, P_{t-1}), \\
    G_t &= \mathrm{T2IM}(P_t), \\
    R_t &= \mathrm{MLLM}_{\text{infer}}(G_t),
\end{align}
where $P_t$ is the textual prompt, $G_t$ the synthesized image, and $R_t$ the inferred idiom (top-1 string).

\vspace{-0.8em}
\subsection{Evaluation}
\vspace{-0.4em}
An LLM judge tests semantic equivalence (with canonicalization of idiom form):
\begin{equation}
    M_t=\mathrm{LLM}_{\text{eval}}(R_t, I_{\text{input}})\in\{\text{true},\text{false}\}.
\end{equation}
The control signal is
\begin{equation}
    C_t=\begin{cases}
        \text{STOP}, & M_t=\text{true}\ \text{or}\ t\geq 5,\\
        \text{CONTINUE}, & \text{otherwise.}
    \end{cases}
\end{equation}

\vspace{-0.8em}
\subsection{Update}
\vspace{-0.4em}
If $C_t=\text{CONTINUE}$, we produce targeted refinements for the next iteration:
\begin{equation}
    U_t=\mathrm{MLLM}_{\text{update}}(R_t, G_t, I_{\text{input}}),
\end{equation}
where $U_t$ lists concrete edits to $P_t$ (e.g., missing objects, composition, emphasis).

\begin{table*}[t]
    \centering
    \small
    \setlength{\tabcolsep}{0.8em}
    \setlength\dashlinedash{1.5pt}
    \setlength\dashlinegap{1.5pt}
    \caption{\small Idiom recognition accuracy (\%) for the iterative pipeline across 10 MLLMs (rows) and 10 LLMs (columns) on 1,000 idioms; the bottom row reports column-wise averages.}
    \label{maintable}
    \begin{tabular}{l|l|ccccc|ccccc}
        \hline
        \multicolumn{2}{c|}{\multirow{2}{*}{\textbf{MLLMs}}} 
        & \multicolumn{5}{c|}{\textbf{Close-sourced LLMs}} 
        & \multicolumn{5}{c}{\textbf{Open-sourced LLMs}} \\
        \cline{3-12}
        \multicolumn{2}{c|}{} & \textbf{GPT} & \textbf{Gemini} & \textbf{Claude} & \textbf{Grok} & \textbf{Doubao} & \textbf{DeepSeek} & \textbf{GPT-OSS}
        & \textbf{Llama} & \textbf{GLM-4.5} & \textbf{Qwen3} \\
        \hline
        \multirow{5}{*}{\rotatebox{90}{\textbf{Close-Sourced}}} 
        & GPT       & 76.9 & 73.7 & 79.8 & 69.5 & 67.3 & 70.1 & 71.1 & 64.8 & 65.9 & 68.7 \\
        & Gemini    & 71.8 & 69.5 & 74.8 & 65.1 & 63.1 & 65.7 & 66.7 & 60.8 & 61.6 & 64.4 \\
        & Claude    & 59.7 & 57.6 & 61.6 & 54.4 & 52.6 & 54.9 & 55.5 & 50.8 & 51.6 & 53.8 \\
        & Grok      & 58.8 & 56.5 & 61.8 & 52.9 & 51.6 & 53.6 & 54.5 & 49.6 & 50.2 & 52.5 \\
        & Doubao    & 55.6 & 53.4 & 58.2 & 50.4 & 49.5 & 51.0 & 51.6 & 47.1 & 48.0 & 50.0 \\
        \hdashline
        \multirow{5}{*}{\rotatebox{90}{\textbf{Open-Sourced}}} 
        & Llama     & 45.9 & 44.1 & 47.8 & 41.6 & 40.3 & 42.1 & 42.6 & 38.8 & 39.5 & 41.2 \\
        & GLM-4.5V  & 48.7 & 46.8 & 50.7 & 44.2 & 42.8 & 44.5 & 45.2 & 41.2 & 41.9 & 43.7 \\
        & Qwen2.5   & 51.3 & 49.3 & 53.4 & 46.5 & 45.1 & 46.9 & 47.6 & 43.4 & 44.1 & 46.0 \\
        & Gemma     & 56.1 & 54.0 & 58.1 & 50.8 & 49.2 & 51.3 & 52.0 & 47.4 & 47.9 & 50.4 \\
        & Mistral   & 28.2 & 27.1 & 29.4 & 25.6 & 24.8 & 25.9 & 26.2 & 24.1 & 24.3 & 25.3 \\
        \hline
        \multicolumn{2}{c|}{Average} & 55.3 & 53.2 & 57.6 & 50.1 & 48.6 & 50.6 & 51.3 & 46.8 & 47.5 & 49.6 \\
        \hline 
    \end{tabular}
\vspace{-2em}
\end{table*}

\begin{table}[t]
\setlength\dashlinedash{1.5pt}
\setlength\dashlinegap{1.5pt}
\caption{\small Overview of evaluated models by provider and release date; bold marks models used in both LLM and MLLM tasks.
}\label{tab:llm_overview}
\vspace{-2em}
\begin{center}
\begin{tabular}{lcc}
\hline
\textbf{Models} & \textbf{Company} & \textbf{Released Date} \\
\hline
\multicolumn{3}{c}{Close Sourced (M)LLMs} \\
\hdashline
\textbf{GPT-5}\cite{gpt51}& OpenAI & August 7, 2025 \\
\textbf{Gemini-2.5-flash}\cite{gem} & Google & June 17, 2025 \\
\textbf{Claude-Sonnet-4}\cite{claude} & Anthropic & May 22, 2025 \\
\textbf{Grok-4}\cite{grok4m} & xAI & July 9, 2025 \\
\textbf{Doubao-Seed-1.6}\cite{doubaom} & ByteDance & June 11, 2025 \\
\hline
\multicolumn{3}{c}{Open Sourced (M)LLMs} \\
\hdashline
\textbf{Llama-4-Maverick} \cite{llama-4-maverick}& Meta & April 5, 2025 \\
DeepSeek-V3.1 \cite{deep}& DeepSeek &  August 21, 2025 \\
GPT-OSS-120b\cite{OSS} & OpenAI &  August 5, 2025 \\
GLM-4.5 \cite{GLM}& Z.ai  & July 28, 2025 \\
Qwen3-32B\cite{qwen3} & Alibaba &  April 29, 2025 \\
GLM-4.5V \cite{glm-4.5v}& Z.ai  & August 11, 2025 \\
Qwen2.5-VL-32B \cite{qwen2} & Alibaba & February 28, 2025 \\
Gemma-3-27b \cite{gemma}& Google & March 12, 2025 \\
Mistral-Small-3.2\cite{mistral-small-3.2} & Mistralai & June 10, 2025 \\
\hline
\multicolumn{3}{c}{Text-to-Image Model} \\
\hdashline
Qwen-Image\cite{qwenimage} & Alibaba & August 4, 2025 \\
\hline
\end{tabular}
\end{center}
    \vspace{-2em}
\end{table}

\vspace{-1em}    
\section{Experiments}
\label{sec:experiments}
\vspace{-1em}
\subsection{Experimental Setup}\label{sec3.1}

\vspace{-0.5em}
Table~\ref{tab:llm_overview} lists the evaluated models: 10 LLMs and 10 MLLMs, with 6 models used in both roles (prompt generation and visual understanding). Closed-source models were accessed via Poe API\footnote{https://poe.com/}; open-source models were accessed via the DeepInfra API\footnote{https://deepinfra.com/}.

We use a single T2IM, Qwen-Image~\cite{qwenimage}, with fixed settings, accessed through Poe API. Images were generated at a fixed resolution of $1024\times1024$. As evidenced in Sec.~\ref{sec3.4}, when prompts are sufficiently detailed, different T2IMs exhibit comparable semantic fidelity. Therefore, we fix the T2IM to Qwen-Image to isolate LLM/MLLM effects and reduce variance and cost.

All experiments use 1{,}000 English idioms. The pipeline runs up to 5 iterations per idiom, generating one image per iteration, and stops early when the MLLM's top-1 inferred idiom matches the target after canonicalization. Accuracy is the proportion of idioms recognized under this criterion. All prompts, hyperparameters, and other settings are unified across LLMs and MLLMs; for exact values and scripts, please refer to the released code, prompts, and images.

\begin{table*}[!t]
\centering
\small
\setlength{\tabcolsep}{1em}
\caption{\small Ablation study showing incremental accuracy improvements from baseline T2IM generation through LLM-augmented prompting and iterative refinements (Claude as fixed LLM).}
\label{tab:ablation}
\begin{tabular}{l|ccccc}
\hline
\multicolumn{1}{c|}{\multirow{2}{*}{\textbf{Configuration}}}
& \multicolumn{5}{c}{\textbf{Close-sourced MLLMs}}\\
\cline{2-6}
& \textbf{GPT} & \textbf{Gemini} & \textbf{Claude} & \textbf{Grok} & \textbf{Doubao}\\
\hline
T2IM & 52.3 & 45.2 & 37.5 & 36.8 & 32.3\\
+LLM & 67.6 (+15.3) & 59.7 (+14.5) & 49.5 (+12.0) & 49.8 (+13.0) & 45.5 (+13.2)\\
Ours(1 update) & 76.2 (+8.6) & 68.8 (+9.1) & 57.5 (+8.0) & 58.3 (+8.5) & 54.3 (+8.8)\\
Ours(2 updates) & 79.3 (+3.1) & 74.5 (+5.7) & 61.0 (+3.5) & 61.5 (+3.2) & 57.8 (+3.5)\\
Ours(3 updates) & 79.7 (+0.4) & 74.8 (+0.3) & 61.5 (+0.5) & 61.8 (+0.3) & 58.2 (+0.4)\\
Ours(4 updates) & 79.8 (+0.1) & 74.8 (+0.0) & 61.6 (+0.1) & 61.8 (+0.0) & 58.2 (+0.0)\\
\hline
\multicolumn{1}{c|}{\multirow{2}{*}{\textbf{Configuration}}}
& \multicolumn{5}{c}{\textbf{Open-sourced MLLMs}}\\
\cline{2-6}
& \textbf{Llama} & \textbf{GLM-4.5V} & \textbf{Qwen2.5} & \textbf{Gemma} & \textbf{Mistral}\\
\hline
T2IM & 28.3 & 26.3 & 30.2 & 29.3 & 16.2\\
+LLM & 38.3 (+10.0) & 38.8 (+12.5) & 42.3 (+12.1) & 44.1 (+14.8) & 23.5 (+7.3)\\
Ours(1 update) & 44.8 (+6.5) & 47.0 (+8.2) & 49.8 (+7.5) & 53.6 (+9.5) & 27.5 (+4.0)\\
Ours(2 updates) & 47.6 (+2.8) & 50.5 (+3.5) & 53.0 (+3.2) & 57.6 (+4.0) & 29.3 (+1.8)\\
Ours(3 updates) & 47.8 (+0.2) & 50.7 (+0.2) & 53.4 (+0.4) & 58.0 (+0.4) & 29.4 (+0.1)\\
Ours(4 updates) & 47.8 (+0.0) & 50.7 (+0.0) & 53.4 (+0.0) & 58.1 (+0.1) & 29.4 (+0.0)\\
\hline
\end{tabular}
\vspace{-0.5em}
\end{table*}

\begin{figure*}[!t]
    \centering
    \includegraphics[width=0.99\linewidth]{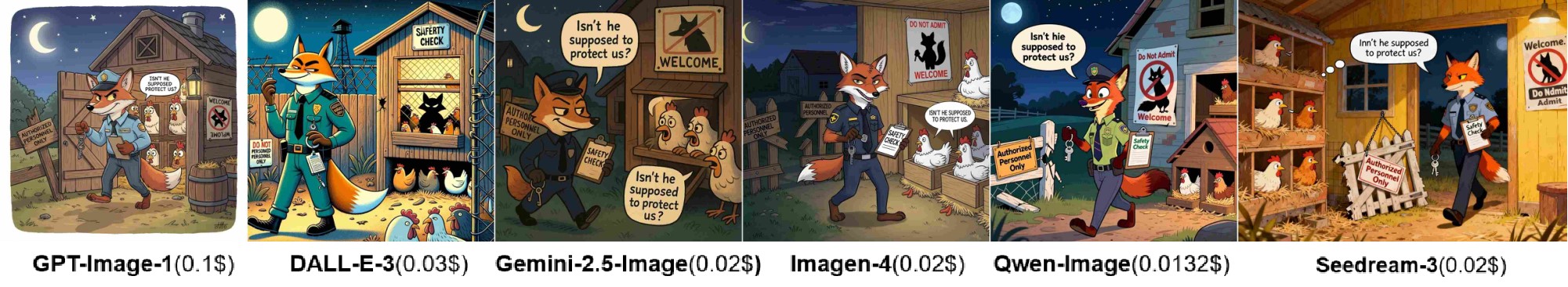}
    \vspace{-1em}
    \caption{\small Comparative visualization of the idiom "fox in a henhouse" generated by different T2IMs using identical prompting. All models were given the prompt:"\textit{Cartoon night scene at a small farm:a sly,sharp-eyed fox in asecurity guard uniform holds a keyring and clipboard labeled"Safety Check,"confidently strolling into a cozy chicken coop.Nervous hens peek from nesting boxes,one whispering to another in a speech bubble,"Isn't he supposed to protect us?"}"}\label{casestudy}
    \label{fig:3}
    \vspace{-1.5em}
\end{figure*}
	
	\vspace{-1em}
\subsection{Main Results and Analysis}
\vspace{-0.5em}
We report top-1 idiom recognition accuracy on 1{,}000 idioms (up to 5 iterations; fixed T2IM). Table~\ref{maintable} varies MLLMs by rows and LLMs by columns; the bottom row is the column-wise average across MLLMs.

MLLM choice dominates performance. GPT MLLM is best across all LLM partners (64.8--79.8\%), followed by Gemini (60.8--74.8\%). Claude and Grok form a mid-tier (mid-50s to low-60s), with Doubao slightly lower (high-40s to high-50s). Among open-source MLLMs, Gemma leads (47.4--58.1\%), approaching closed-source Doubao on several columns; Qwen2.5 and GLM-4.5V cluster in the mid-40s to low-50s, Llama in the low-40s, and Mistral trails (24--29\%). The best--worst MLLM gap at a fixed LLM can exceed 50 points (e.g., with Claude as LLM: 79.8 vs.\ 29.4), underscoring that visual understanding is the primary variance source; the MLLM ordering is largely stable across LLM partners.

LLM effects are smaller but non-negligible. Column-wise averages span 46.8--57.6\%, with Claude strongest for prompt generation (57.6\%), followed by GPT (55.3\%) and Gemini (53.2\%). While careful linguistic prompting helps, the narrower column range versus row spreads indicates the chief bottleneck lies in MLLM visual reasoning. The best observed pairing is GPT (MLLM) with Claude (LLM) at 79.8\%.

Open-source alternatives are competitive under cost constraints. As MLLMs, Gemma (up to 58.1\%) nears closed-source Doubao; as LLMs, GPT-OSS averages 51.3\%, exceeding Doubao (48.6\%) and narrowing the gap to top closed-source LLMs.
\vspace{-1em}
\subsection{Ablation Study}\label{ablation}
\vspace{-0.3em}
We ablate three configurations on 1{,}000 idioms using the same T2IM (Qwen-Image, $1024{\times}1024$) and protocol as Sec.~\ref{sec3.1}; accuracy is top-1 idiom match after canonicalization with early stopping (max 5 iterations).

\textbf{Configurations.} (i) \emph{T2IM-only}: the idiom string is used directly as the prompt; no LLM is involved, and the MLLMs performs recognition. (ii) \emph{+LLM}: one-shot LLM-generated prompt (Claude fixed for prompting); no updates. (iii) \emph{Ours ($k$ updates)}: iterative prompt refinement with $k{=}1\ldots4$; Claude produces edits each round; the MLLMs handles recognition. All other settings are fixed; only the presence/number of updates varies.

\textbf{Results (Table 3).} T2IM-only yields 16.2–52.3\% across MLLMs, indicating direct generation under-specifies figurative intent. Adding an LLM prompt improves accuracy by +7.3–+15.3 points on every MLLM, showing the benefit of linguistic decomposition. Iterative refinement adds gains: the first update contributes +4.0–+9.5 points; later updates have diminishing returns, with negligible improvements by the 4th update, indicating convergence within 3–4 iterations.

\textbf{Takeaway.} With the T2IM held constant, performance gains are primarily attributable to (a) introducing linguistic guidance and (b) a small number of targeted updates, beyond which additional iterations bring little benefit.

\vspace{-1em}
\subsection{Case Study}\label{sec3.4}
\vspace{-0.5em}

Using an identical LLM-crafted prompt for ``fox in a henhouse'' (Fig.~\ref{fig:3}; $1024\times 1024$), several state-of-the-art T2IMs produced images that our MLLM consistently mapped to the target idiom despite stylistic differences. We observed the same pattern on 50 additional idioms across 6 T2IMs; for brevity, we release the selection protocol, idiom list, prompts, and images in the repository. Together with the ablations in Sec.~\ref{ablation}—which attribute most gains to LLM guidance and a small number of iterative updates—this case study suggests that, under detailed prompting, T2IM choice is secondary. Accordingly, we fix Qwen-Image as the T2IM for all experiments.

\vspace{-2em}
\section{Conclusion}
\label{sec:conclusion}
\vspace{-0.5em}

We introduced an iterative framework for generating idiom-based visual puns and released a large-scale dataset of 1{,}000 idioms with paired prompts and images for benchmarking multimodal generation and understanding; resources are publicly available as cited in the abstract. Experiments varying 10 LLMs and 10 MLLMs with a fixed T2IM show that MLLM choice is the principal driver of recognition accuracy, while LLMs have smaller but consistent effects; most gains arise within 2--3 refinement rounds. Limitations include reliance on a single T2IM configuration and automatic, MLLM-based evaluation; future work will expand T2IM diversity and incorporate human studies and cross-lingual idioms. We hope these resources support more reliable assessment and progress in creative multimodal reasoning.

\label{sec:refs}
{ 
    \small
    \begin{spacing}{0.8}
        \bibliographystyle{IEEEbib}
        \bibliography{inproceedings,article}
    \end{spacing}
}
	
\end{document}